\documentclass[11pt]{article}
\usepackage{eacl2017}
\usepackage{times}
\usepackage{url}
\usepackage{latexsym}
\usepackage{graphicx}
\usepackage{color}
\usepackage{amssymb}
\usepackage{amsmath}
\eaclfinalcopy 

\def\eaclpaperid{527} 

\newcommand{\whethermath}[1]{\ifmmode{#1}\else{$#1$}\fi}
\newcommand{\uprm}[1]{\whethermath{^{\mbox{\scriptsize #1}}}}

\title{A Copy-Augmented Sequence-to-Sequence Architecture Gives \\Good Performance on Task-Oriented Dialogue}

\author{Mihail Eric \and Christopher D. Manning \\
  Computer Science Department\\
  Stanford University\\
  {\tt meric@cs.stanford.edu}, {\tt manning@stanford.edu} \\}

\date{}

\begin{document}
\maketitle
\begin{abstract}
  Task-oriented dialogue focuses on conversational agents that participate in dialogues with user goals on domain-specific topics. In contrast to chatbots, which simply seek to sustain open-ended meaningful discourse, existing task-oriented agents usually explicitly model user intent and belief states. This paper examines bypassing such an explicit representation by depending on a latent neural embedding of state and learning selective attention to dialogue history together with copying to  incorporate relevant prior context. We complement recent work by showing the effectiveness of simple sequence-to-sequence neural architectures with a copy mechanism. Our model outperforms more complex memory-augmented models by 7\% in per-response generation and is on par with the current state-of-the-art on DSTC2, a real-world task-oriented dialogue dataset.
\end{abstract}

\section{Introduction}

  Effective task-oriented dialogue systems are becoming important as society progresses toward using voice for interacting with devices and performing everyday tasks such as scheduling.
  To that end, research efforts have focused on using machine learning methods to train agents using dialogue corpora. One line of work has tackled the problem using partially observable Markov decision processes and reinforcement learning with carefully designed action spaces~\cite{Young:13}. However, the large, hand-designed action and state spaces make this class of models brittle and unscalable, and in practice most deployed dialogue systems remain hand-written, rule-based systems.

    Recently, neural network models have achieved success on a variety of natural language processing tasks ~\cite{Bahdanau:14,NIPS2014_5346,NIPS2015_5635}, due to their ability to implicitly learn powerful distributed representations from data in an end-to-end trainable fashion. This paper extends recent work examining the utility of distributed state representations for task-oriented dialogue agents, without providing rules or manually tuning features. 

  One prominent line of recent neural dialogue work has continued to build systems with modularly-connected representation, belief state, and generation components~\cite{Wen:16}. These models must learn to explicitly represent user intent through intermediate supervision, and hence suffer from not being truly end-to-end trainable. Other work stores dialogue context in a memory module and repeatedly queries and reasons about this context to select an adequate system response~\cite{Bordes:16}. While reasoning over memory is appealing, these models simply choose among a set of utterances rather than generating text and also must have temporal dialogue features explicitly encoded.

  However, the present literature lacks results for now standard sequence-to-sequence architectures, and we aim to fill this gap by building increasingly complex models of text generation, starting with a vanilla sequence-to-sequence recurrent architecture. The result is a simple, intuitive, and highly competitive model, which outperforms the more complex model of \newcite{Bordes:16} by 6.9\%. Our contributions are as follows: 1) We perform a systematic, empirical analysis of increasingly complex sequence-to-sequence models for task-oriented dialogue, and 
2) we develop a recurrent neural dialogue architecture augmented with an attention-based copy mechanism that is able to significantly outperform more complex models on a variety of metrics on realistic data.

\section{Architecture}

  We use neural encoder-decoder architectures 
to frame dialogue as a sequence-to-sequence learning problem. Given a dialogue between a user (\emph{u}) and a system (\emph{s}), we represent the dialogue utterances as $\{(u_1, s_1), (u_2, s_2), \ldots ,(u_k, s_k) \}$ where $k$ denotes the number of turns in the dialogue. At the $i\uprm{th}$ turn of the dialogue,
we encode the aggregated dialogue context composed of the tokens of $(u_1, s_1, \ldots, s_{i-1}, u_i)$. Letting $x_1, \ldots  , x_m$ denote these tokens,
we first embed these tokens using a trained embedding function $\phi^{emb}$ that maps each token to a fixed-dimensional vector. These mappings are fed into the encoder to produce context-sensitive hidden representations $h_1, \ldots , h_m$. 

  The vanilla Seq2Seq decoder predicts the tokens of the $i\uprm{th}$ system response $s_i$ by first computing decoder hidden states via the recurrent unit. We denote $\tilde h_1, \ldots, \tilde h_n$ as the hidden states of the decoder and $y_1, \ldots, y_n$ as the output tokens.  We extend this decoder with an attention-based model \cite{Bahdanau:14,Luong:15a}, where, at every time step $t$ of the decoding, an attention score $a_{i}^t$ is computed for each hidden state $h_i$ of the encoder, using the attention mechanism of ~\cite{NIPS2015_5635}. Formally this attention can be described by the following equations:
      \begin{align}
          u_i^t &= v^T\tanh(W_1h_i + W_2\tilde h_t) \\
          a_i^t &= \textrm{Softmax}(u_i^t)\\
          \tilde h_t' &= \displaystyle \sum_{i=1}^m a_i^th_i \\
          o_t &= U[\tilde h_t, \tilde h_t'] \\
          y_t &= \textrm{Softmax}(o_t)
      \end{align}

  \noindent where $W_1$, $W_2$, $U$, and $v$ are trainable parameters of the model and $o_t$ represents the logits over the tokens of the output vocabulary $V$. During training, the next token $y_t$ is predicted so as to maximize the log-likelihood of the correct output sequence given the input sequence.

  An effective task-oriented dialogue system must have powerful language modelling capabilities and be able to pick up on relevant entities of an underlying knowledge base. One source of relevant entities is that they will commonly have been mentioned in the prior discourse context. Recent literature has shown that incorporating a copying mechanism into neural architectures improves performance on various sequence-to-sequence tasks including code generation, machine translation, and text summarization ~\cite{gu-EtAl:2016:P16-1,ling-EtAl:2016:P16-1,gulcehre-EtAl:2016:P16-1}. We therefore augment the attention encoder-decoder model with an attention-based copy mechanism in the style of ~\cite{jia-liang:2016:P16-1}. In this scheme, during decoding we compute our new logits vector as $o_t = U[\tilde h_t, \tilde h_t', a^t_{[1:m]}]$ where $a^t_{[1:m]}$ is the concatenated attention scores of the encoder hidden states, and we are now predicting over a vocabulary of size $|V|+m$. The model, thus, either predicts a token $y_t$ from $V$ or copies a token $x_i$ from the encoder input context, via the attention score $a_{i}^t$. 
  Rather than copy over any token mentioned in the encoder dialogue context, our model is trained to only copy over entities of the knowledge base mentioned in the dialogue context, as this provides a conceptually intuitive goal for the model's predictive learning: as training progresses it will learn to either predict a token from the standard vocabulary of the language model thereby ensuring well-formed natural language utterances, or to copy over the relevant entities from the input context, thereby learning to extract important dialogue context.

  In our best performing model, we augment the inputs to the encoder by adding entity type features. Classes present in the knowledge base of the dataset, namely the 8 distinct entity types referred to in Table 1, are encoded as one-hot vectors. Whenever a token of a certain entity type is seen during encoding, we append the appropriate one-hot vector to the token's word embedding before it is fed into the recurrent cell. These type features improve generalization to novel entities by allowing the model to hone in on positions with particularly relevant bits of dialogue context during its soft attention and copying. Other cited work using the DSTC2 dataset ~\cite{Bordes:16,Liu-Perez:16,Seo:16} implement similar mechanisms whereby they expand the feature representations of candidate system responses based on whether there is lexical entity class matching with provided dialogue context. In these works, such features are referred to as \emph{match} features.

  All of our architectures use an LSTM cell as the recurrent unit ~\cite{Hochreiter:97} with a bias of 1 added to the forget gate in the style of ~\cite{Pham:14}.

\section{Experiments}
 
\subsection{Data}
  For our experiments, we used dialogues extracted from the Dialogue State Tracking Challenge 2 (DSTC2)~\cite{Henderson:14}, a restaurant reservation system dataset. While the goal of the original challenge was building a system for inferring dialogue state, for our study, we use the version of the data from \newcite{Bordes:16}, which ignores the dialogue state annotations, using only the raw text of the dialogues. The raw text includes user and system utterances as well as the API calls the system would make to the underlying KB in response to the user's queries. Our model then aims to predict both these system utterances and API calls, each of which is regarded as a turn of the dialogue. We use the train/validation/test splits from this modified version of the dataset. The dataset is appealing for a number of reasons: 1) It is derived from a real-world system so it presents the kind of linguistic diversity and conversational abilities we would hope for in an effective dialogue agent. 2) It is grounded via an underlying knowledge base of restaurant entities and their attributes. 3) Previous results have been reported on it so we can directly compare our model performance. We include statistics of the dataset in Table 1.

  \subsection{Training}

    We trained using a cross-entropy loss and the Adam optimizer ~\cite{Kingma:15}, applying dropout ~\cite{Hinton:12} as a regularizer to the input and output of the LSTM.  We identified hyperparameters by random search, evaluating on a held-out validation subset of the data. Dropout keep rates ranged from 0.75 to 0.95. We used word embeddings with size 300, and hidden layer and cell sizes were set to 353, identified through our search.  We applied gradient clipping with a clip-value of 10 to avoid gradient explosions during training.
    The attention, output parameters, word embeddings, and LSTM weights were randomly initialized from a uniform unit-scaled distribution in the style of ~\cite{Sussillo:15}.

\begin{table}
\centering
\small
\begin{tabular}{cc}
\begin{tabular}{|l|l|}
\hline
Avg. \# of Utterances Per Dialogue & 14 \\
Vocabulary Size & 1,229 \\
Training Dialogues & 1,618  \\
Validation Dialogues & 500  \\
Test Dialogues & 1,117  \\
\# of Distinct Entities & 452 \\
\# of Entity (or Slot) Types & 8 \\
\hline
\end{tabular}
\end{tabular}
\caption{Statistics of DSTC2}\label{tab:accents}
\end{table}

\subsection{Metrics}
  Evaluation of dialogue systems is known to be difficult ~\cite{liu-EtAl:2016:EMNLP20163}. We employ several metrics for assessing specific aspects of our model, drawn from previous work:
  \begin{itemize}
    \item \textbf{Per-Response Accuracy}: 
Bordes and Weston ~\shortcite{Bordes:16} report a per-turn response accuracy, which tests their model's ability to select the system response at a certain timestep. Their system does a multiclass classification over a predefined candidate set of responses, which was created by aggregating all system responses seen in the training, validation, and test sets. Our model actually generates each individual token of the response, and we consider a prediction to be correct only if every token of the model output matches the corresponding token in the gold response. Evaluating using this metric on our model is therefore significantly more stringent a test than for the model of Bordes and Weston ~\shortcite{Bordes:16}.
    \item \textbf{Per-Dialogue Accuracy}: Bordes and Weston \shortcite{Bordes:16}  also report a per-dialogue accuracy, which assesses their model's ability to produce every system response of the dialogue correctly. We calculate a similar value of dialogue accuracy, though again our model generates every token of every response. 
    \item \textbf{BLEU}: We use the BLEU metric, commonly employed in evaluating machine translation systems ~\cite{papineni-EtAl:2002:ACL}, which has also been used in past literature for evaluating dialogue systems ~\cite{Ritter:11a,li-EtAl:2016:N16-11}. We calculate average BLEU score over all responses generated by the system, and primarily report these scores to gauge our model's ability to accurately generate the language patterns seen in DSTC2.
    \item \textbf{Entity F$_1$}: Each system response in the test data defines a gold set of entities. To compute an entity F$_1$, we micro-average over the entire set of system dialogue responses. This metric evaluates the model's ability to generate relevant entities from the underlying knowledge base and to capture the semantics of the user-initiated dialogue flow.
  \end{itemize}
 
  Our experiments show that sometimes our model generates a response to a given input that is perfectly reasonable, but is penalized because our evaluation metrics involve direct comparison to the gold system output. For example, given a user request for an \emph{australian restaurant}, the gold system output is \emph{you are looking for an australian restaurant right?} whereas our system outputs \emph{what part of town do you have in mind?}, which is a more directed follow-up intended to narrow down the search space of candidate restaurants the system should propose. This issue, which recurs with evaluation of dialogue or other generative systems, could be alleviated through more forgiving evaluation procedures based on beam search decoding.

\subsection{Results}
In Table 2, we present the results of our models compared to the reported performance of the best performing model  of ~\cite{Bordes:16}, which is a variant of an end-to-end memory network ~\cite{NIPS2015_5846}. Their model is referred to as \emph{MemNN}. We also include the model of ~\cite{Liu-Perez:16}, referred to as \emph{GMemNN}, and the model of ~\cite{Seo:16}, referred to as \emph{QRN}, which currently is the state-of-the-art. 
In the table, Seq2Seq refers to our vanilla encoder-decoder architecture with (1), (2), and (3) LSTM layers respectively. +Attn refers to a 1-layer Seq2Seq with attention-based decoding. +Copy refers to +Attn with our copy-mechanism added. +EntType refers to +Copy with entity class features added to encoder inputs.

We see that a 1-layer vanilla encoder-decoder is already able to significantly outperform \emph{MemNN} in both per-response and per-dialogue accuracies, despite our more stringent setting. Adding layers to Seq2Seq leads to a drop in performance, suggesting an overly powerful model for the small dataset size. Adding an attention-based decoding to the vanilla model increases BLEU although per-response and per-dialogue accuracies suffer a bit. Adding our attention-based entity copy mechanism achieves substantial increases in per-response accuracies and entity F$_1$. Adding entity class features to +Copy achieves our best-performing model, in terms of per-response accuracy and entity F$_1$. This model achieves a 6.9\% increase in per-response accuracy on DSTC2 over \emph{MemNN}, including +1.5\% per-dialogue accuracy, and is on par with the performance of \emph{GMemNN}, including beating its per-dialogue accuracy. It also achieves the highest entity F$_1$.

\begin{table}
\centering
\small
\begin{tabular}{@{}llcccc@{}}
{\bf Data} & {\bf Model} & {\bf Per-} & {\bf Per} & {\bf BLEU} & {\bf Ent.} \\
 &  & {\bf Resp.} & {\bf Dial.} & & {\bf F$_1$}\\
\hline
{\bf Test} & \emph{MemNN} & 41.1 & 0.0 & -- & -- \\
{\bf set} & \emph{GMemNN} & 48.7 & 1.4 & -- & -- \\
& \emph{QRN} & 50.7 & -- & -- & -- \\
& Seq2Seq (1) & 46.4 & 1.5 & 55.0 & 69.7 \\
& Seq2Seq (2) & 43.5 & 1.3 & 54.2 & 67.3 \\
& Seq2Seq (3) & 44.2 & \textbf{1.7} & 55.4 & 65.9 \\
& \hspace*{1em} + Attn. & 46.0 & 1.4 & \textbf{56.6} & 67.1 \\
& \hspace*{1em} + Copy & 47.3 & 1.3 & 55.4 & 71.6 \\ 
& \hspace*{1em} + EntType & \textbf{48.0} & 1.5 & 56.0 & \textbf{72.9} \\
\hline
{\bf Dev} & Seq2Seq (1) & 57.0 & 3.6 & 72.1 & 68.7 \\
{\bf set} & Seq2Seq (2) & 54.1 & 3.0 & 71.3 & 66.3 \\
& Seq2Seq (3) & 54.0 & 3.2 & 71.5 & 64.3 \\
& \hspace*{1em} + Attn. & 55.2 & 3.4 & 71.9 &  66.1\\
& \hspace*{1em} + Copy & 58.9 & 3.6 & 73.1 & 72.5 \\
& \hspace*{1em} + EntType & 59.2 & 3.4 & 72.7 & 72.3 \\
\hline
\end{tabular}
\caption{Evaluation on DSTC2 test (top) and dev (bottom) data. Bold values indicate our best performance. A dash indicates unavailable values.}\label{tab:accents}
\end{table}

\section{Discussion and Conclusion}
  
    We have iteratively built out a class of neural models for task-oriented dialogue that is able to outperform other more intricately designed neural architectures on a number of metrics. The model incorporates in a simple way abilities that we believe are essential to building good task-oriented dialogue agents, namely maintaining dialogue state and being able to extract and use relevant entities in its responses, without requiring intermediate supervision of dialogue state or belief tracker modules. Other dialogue models tested on DSTC2 that are more performant in per-response accuracy are equipped with sufficiently more complex mechanisms than our model. Taking inspiration from ~\cite{NIPS2015_5846} and ~\cite{Srivastava:15}, \emph{GMemNN} uses an explicit memory module as well as an adaptive gating mechanism to learn to attend to relevant memories. The \emph{QRN} model employs a variant of a recurrent unit that is intended to handle local and global interactions in sequential data. We contrast with these works by bootstrapping off of more empirically accepted Seq2Seq architectures through intuitive extensions, while still producing highly competitive models.

    We attribute the large gains in per-response accuracy and entity F$_1$ demonstrated by our +EntType to its ability to pick out the relevant KB entities from the dialogue context fed into the encoder. In Figure 1, we see the attention-based copy weights of the model, indicating that the model is able to learn the relevant entities it should focus on in the input context. The powerful language modelling abilities of the Seq2Seq backbone allow smooth integration of these extracted entities into both system-generated API calls and natural language responses as shown in the figure. 

  \begin{table}
\centering
\small
\begin{tabular}{cc}
\begin{tabular}{|l|}
\hline
cheap restaurant in east part of town  \\
\emph{api\_call r\_cuisine east cheap} \\
$<$silence$>$ \\
\emph{the\_missing\_sock is a nice place in the} \\
\emph{ east of town and the prices are cheap} \\
address \\
\emph{sure, the\_missing\_sock is on the\_missing\_sock\_address} \\
phone number \\
\emph{the phone number of the\_missing\_sock is }\\
\emph{the\_missing\_sock\_phone} \\
thank you good bye \\
\emph{you are welcome}\\
\hline

\hline
\end{tabular}
\end{tabular}
\caption{Sample dialogue generated. System responses are in italics. The dataset uses fake addresses and phone numbers.}\label{tab:sample}
\end{table}

  \begin{figure}
    \caption{Attention-copy weights for a generated natural language response (top) and API call (bottom). The decoder output is displayed vertically and the encoder input is abbreviated for display.}
    \centering
    \includegraphics[width=0.5\textwidth]{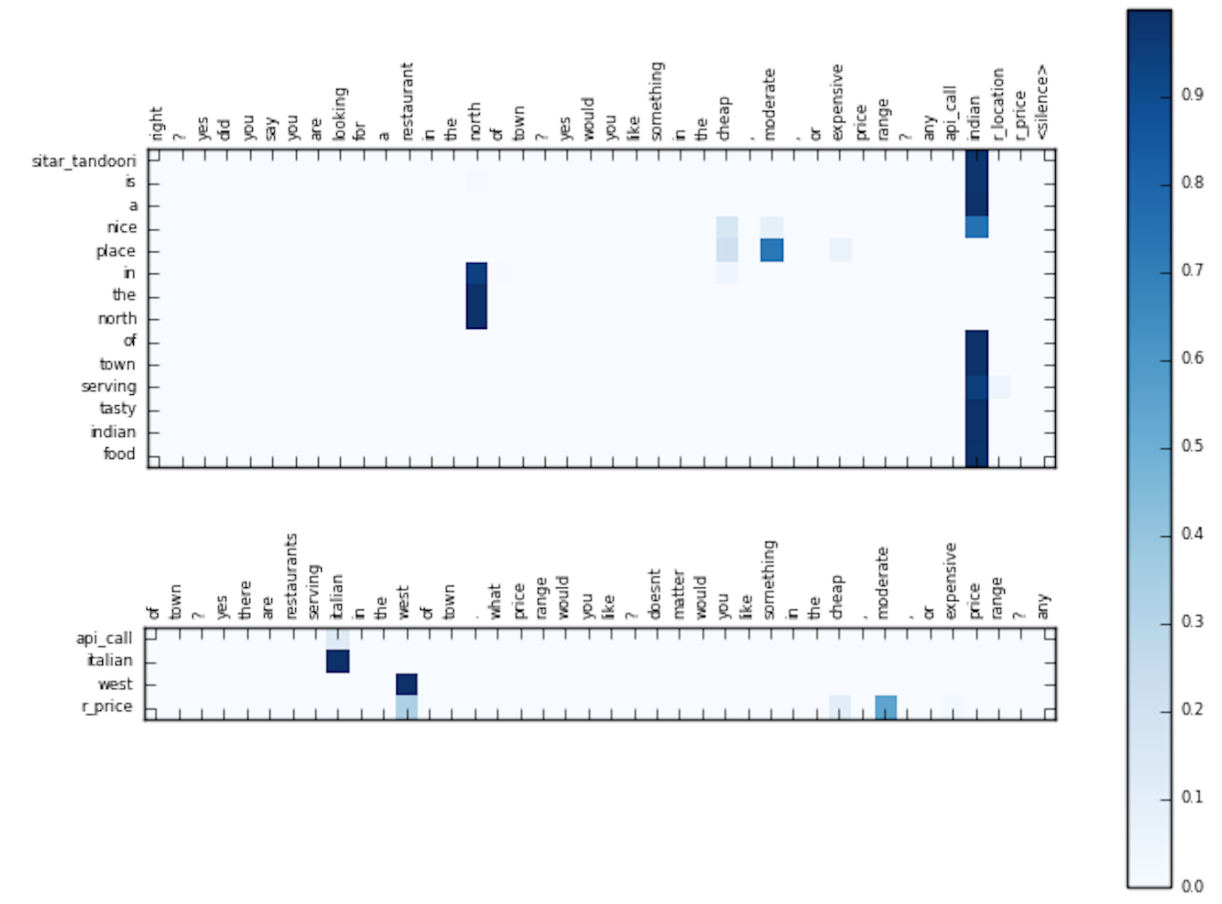}
  \end{figure}

  The appeal of our model comes from the simplicity and effectiveness of framing system response generation as a sequence-to-sequence mapping with a soft copy mechanism over relevant context. Unlike the task-oriented dialogue agents of Wen et. al \shortcite{Wen:16}, our architecture does not explicitly model belief states or KB slot-value trackers, and we preserve full end-to-end-trainability. Further, in contrast to other referenced work on DSTC2, our model offers more linguistic versatility due to its generative nature while still remaining highly competitive and outperforming other models. Of course, this is not to deny the importance of dialogue agents which can more effectively use a knowledge base to answer user requests, and this remains a good avenue for further work.
Nevertheless, we hope this simple and effective architecture can be a strong baseline for future research efforts on task-oriented dialogue.

\section*{Acknowledgments}

The authors wish to thank the reviewers, Lakshmi Krishnan, Francois Charette, and He He for their valuable feedback and insights. We gratefully acknowledge the funding of the Ford Research and Innovation Center, under Grant No. 124344. The views expressed here are those of the authors and do not necessarily represent or reflect the views of the Ford Research and Innovation Center.

\bibliography{meric_eacl2017}
\bibliographystyle{meric_eacl2017}

\appendix

\end{document}